\documentclass[conference]{IEEEtran}
\IEEEoverridecommandlockouts
\usepackage{amsmath,amssymb,amsfonts}
\usepackage{xcolor,float}
\usepackage{booktabs,cite}
\usepackage{amsmath,amssymb,amsfonts}
\usepackage{graphicx,textcomp,url}
\usepackage{hyperref}
\begin{document}
\title{Selective Survey: Most Efficient Models and Solvers for Integrative Multimodal Transport} 
\author{
	\IEEEauthorblockN{Oliviu Matei, Erdei Rudolf, Camelia-M. Pintea}
	\IEEEauthorblockA{\textit{Technical University Cluj-Napoca, Romania} \\
	oliviu.matei@holisun.com,\\rudolf.erdei@holisun.com, dr.camelia.pintea@ieee.org}
}

\maketitle
\begin{abstract}
In the family of Intelligent Transportation Systems (ITS), Multimodal Transport Systems (MMTS) have placed themselves as a mainstream transportation mean of our time as a feasible integrative transportation process. The Global Economy progressed with the help of transportation. The volume of goods and distances covered have doubled in the last ten years, so there is a high demand of an optimized transportation, fast but with low costs, saving resources but also safe, with low or zero emissions. Thus, it is important to have an overview of existing research in this field, to know what was already done and what is to be studied next. The main objective is to explore a beneficent selection of the existing research, methods and information in the field of multimodal transportation research, to identify industry needs and gaps in research and provide context for future research. The selective survey covers multimodal transport design and optimization in terms of: cost, time, and network topology. The multimodal transport theoretical aspects, context and resources are also covering various aspects. The survey's selection includes nowadays best  methods and solvers for Intelligent Transportation Systems (ITS). The gap between theory and real-world applications should be further solved in order to optimize the global multimodal transportation system. 
\end{abstract}\footnote{Article-version; no images included.}
\section{Introduction}\label{sec:introduction}
The current context of World Globalization has raised many difficult problems regarding the transportation of goods. The products are hauled over large distances of land and water, and more often get to travel by more than one means of transport: by ships, planes, trucks (see ~\cite{van2000congestion}); all these lead to the Multimodal Transportation Systems (MMTS).

In contrast with classical, single mean transportation, multi-modal transportation has multiple constraints, for example in ~\cite{litman2017introduction}, different optimization processes as parcel loading, and transfer between transports. In another context, of today's global warming and increased pollution, it's a necessity to also globally lower gas emission. The environmental goal correlated with the economical performance could be reached through several ways including an optimized transport planning and using appropriate resources. 

In~\cite{SteadieSeifi} literature review on Multimodal freight transportation planning, several strategic planning issues within multi-modal freight transportation and tactical planning problems are shown. Complex operational planning for real-time requirements of multimodal operators, carriers and shippers, not previously addressed at strategic and tactical levels, are described.

The main models with related solvers and proposed future research are included. A detailed review, with an analysis of the optimization-based decision-making models for the problem of {\it Disaster Recovery Planning of Transportation Networks (DRPTN)} is provided by ~\cite{Zamanifar}. The authors described the phases of optimization-based decision-making models and investigate their methodologies. Nevertheless the authors identify some challenges and opportunities, discussed research improvement and made suggestions for possible future research. 

A recent systematic review about dynamic pricing techniques for {\it Intelligent Transportation System (ITS)} in smart cities was published by~\cite{SaharanBK20}. The authors included existing ITS techniques with pertinent overviews and discussions about problems related to electric vehicles (EVs) used for reducing the peak loads and congestion, respectively increasing mobility. 

The current work overviews the multimodal transport. Section~\ref{sec:whatismultimodal} presents prerequisites related to the multimodal transport and the context around it. Follows the section~\ref{sec:challenges} presenting the characteristics and the challenges related to the transport. Further on, methods of planning (section~\ref{sec:planning}) and optimization (section~\ref{sec:optimization}) in terms of time (section~\ref{sec:timeopt}), cost (section~\ref{sec:costopt}) and network topology (section~\ref{sec:netopt}). Existing unimodal transport models and solver with possible future extension to multimodal features are included in section~\ref{sec:theoretical}. Section~\ref{sec:conclusions} draws the important conclusions about multimodal transport.  

\medskip

\section{What is Multimodal Transportation?}\label{sec:whatismultimodal}
The multimodal transport is defined by the UN Convention on International Multimodal Transport of Goods as follows. 

\medskip

\noindent{\bf Definition}\cite{peplowska2019codification}\label{clas}
The {\it multimodal transport} is the transport of goods from one place to another, usually located in a different country, by at least two means of transportation. 

\medskip

\noindent{\bf Mathematical formalization.} The transportation problem was first formalized by~\cite{monge} and extended by~\cite{kantor}. Today there are mathematical optimization techniques as for example Newton, Quasi-Newton methods and Gauss-Newton techniques already used or to be further used in relation with transportation e.g. in assignment models to calibrate the traffic and transit; see: ~\cite{karballaeezadeh2020intelligent,ticalempirical,kamel2019integrated}.

\newpage
\noindent{\bf Optimization.} A classical transportation of goods implies direct links and one mode of transportation, a shortest path route, from a sender to a receiver of goods, see ~\cite{zografos2008algorithms}; the  multimodal transport, implies complex links and more than one mode of transportation. Optimization of classical transportation routes is fairly easy and intensely studied topic. State-of-art algorithms already exist, like Dijkstra's Algorithm (see~\cite{jianya1999efficient}), or Clarke-Wright
 technique as in~\cite{golden1977implementing}. These approaches use a single mean of transport, with a single warehouse and one or more clients (or receivers). 

\bigskip

\noindent{\bf Real-life scenarios.} In the complex reality, goods can be transported in any direction, for example inside a country there are couriers delivering from any side of a country to another side; using just trucks to perform a complete task would be impossible, as the problem has $O(n^2)$ complexity for $n$ cities to reach. Optimizing this case meant designing a system with a central warehouse or hub, where all the goods are unloaded, sorted according to the destination and finally loaded on the respective trucks and dispatching them towards the destination. This optimization alone reduces the required number of trucks (the same truck makes a round trip from each city to the central hub)~\cite{zhang2013optimization}.

But what about the {\it larger countries}, or about {\it international transport}? The multimodal transportation has the advantage of moving a huge amount of goods, in the hundreds of thousands of tons at once, via large ships, over very large distances.

\section{Characteristics and challenges of Multimodal Transportation}\label{sec:challenges}
This section focuses on the challenges of the multimodal transportation, both for {\it passengers} (Section~\ref{sec:passengers}) and {\it freight} (Section~\ref{sec:freight}).  The majority of received goods are moved with many transportation modes, e.g. ships, airplanes, trucks. In the first stages of the transport, the sorting hubs aggregate all the goods from different senders, establish their destination and assign them a way of dispatch. Routes may be calculated at this step to assess the most economical ones, both in cost and time. Goods with the same route are grouped and loaded on the same shipment mode. 

Direct {\it consequence}: when reaching the end of a route with each transportation mode there must be a sorting/dispatching hub. These operations of unloading, sorting, grouping and dispatching are repeated at every hub and with a highly time consuming action. As a consequence, the intermediate hubs have to be very organized so as to limit the time spent in that point, and also their number has to be kept low enough. As an disadvantage, the very large number of shipping hubs will dramatically increase the transport cost, due to the number of units of transport used.

\indent Based on these features, several challenges arise: How can we make shipping from A to B cheaper, quicker, and with the least environmental impact? How can we calculate the optimum number of hubs with maximum benefits? What is the optimal way of transporting goods between hubs while avoiding their weaknesses? 
 
\subsection{Models \& Solvers.}

\smallskip

\noindent $>$ {\bf Real-life scenarios.}

\smallskip

\indent -- {\bf HAZMAT: transport Security Vulnerability Assessment (SVA)} by~\cite{reniers2013method}.
The hazardous materials HAZMAT transport SVA assess the relative security risk levels of the different modes of hazardous freight transport models, e.g. road, inland waterways, pipelines or railway. The policymakers could use this tool to assess the user-friendly security in multi-modal transport. The HAZMAT model follows:
\begin{itemize}
\item[-] The routes are split into smaller segments. 
\item[-] The probability scores of security-related risks in which dangerous freight is involved and possibly causing fatalities in the surrounding population, are determined for each segment. 
\item[-] The impact of injury scenarios are computed in terms of the number of people within the 1\% lethal distance of the incident center. 
\item[-] Based on these probability and impact scores, transport route security risk levels are determined.
\item[-] The transshipment risks are considered for determining the final transport route security risk levels.
\end{itemize}
 
The inter-modal risk is determined on the minimum security risk path, considering only the risks of the individual segments of a transport route and include also the number of intermodal transshipment. 

The risk with the transshipment is defined as: $R_t = R_{nt}(1 + x(nts))$ where $R_t$ is the security with the transshipment, $R_{nt}$ is the security risk without transshipment, $x$ is the weight factor for importance of transshipment risks, compared with the transportation risks and $nts$ is the number of transshipment. The model was implemented with CPLEX studio and OPL; it was successfully tested on two multimodal networks with highways and railways.

\smallskip

\indent -- {\bf New Delhi, Indian busy urban area MMTS} by~\cite{kumar2013performance}. In 2021 the Delhi population will be around 23 million, therefore public transit should be integrated. In~\cite{kumar2013performance} MMTS focuses on reducing congestion on roads and improving transfers and interchanges between modes. Delhi`s public transport will grow from a 60\% of the total number of vehicular trips to at least at 80\% in 2021; 15 million trips per day by 2021 in the Integrated Rail-cum-Bus Transit, plus 9 million with other modes are estimated. The Delhi public transport model is illustrated, evaluated and its performance is discussed in~\cite{kumar2013performance}.

\medskip

 The {\bf performance} of the MMTS is quantified using the following measures.

\begin{itemize}
\item[-] {\it Travel Time Ratio (TTR)}: a large TTR value leads to a less competitive public transport, e.g. $TTR\in [1,5]$; 
\item[-] {\it Level of Service (LS)} is a ratio of Out-of-vehicle Travel Time (OVTT) to In-Vehicle Travel Time (IVTT); a large LS measure leads to a less attractive public transport, e.g. $LS\in [1.2,5.0]$; 
\item[-] {\it Inter-connectivity Ratio (IR)} is the ratio of access and egress time to the total trip time; $IR\in [0,1]$;
\item[-] {\it Passenger Waiting Index (PWI)} is the ratio of mean passenger waiting time to transport services' frequency; the number of boarding passengers is less or equal with the  available space in the transport mode; $PWI\in [0,1]$;
\item[-] {\it Running Index (RI)} is the ratio of total service time to total travel time; a large RI leads to a decreased efficiency of the system; $RI\in [0,1]$; \end{itemize}

In particular, for the New Delhi case study: $TTR=1.3$ shows a competitive public transport; LS, the mean $OVTT/IVTT > 1$ thus people spend more time out-of-vehicle than in-vehicle; $IR\in [0.2,0.5]$ value shows that inter-connectivity between transportation modes should be improved; $PWI=0.825$ for the Metro is recommended as the mean passenger waiting time is similar with Metro's frequency; $RI=0.7681$, indicates that passenger satisfaction should be improved. 

\indent -- {\bf ARKTRANS. The Norwegian MMTS framework architecture} by ~\cite{natvig2006arktrans} and \cite{natvig2010flexible}. The framework offers an overview of all the major systems running in Norway that will hopefully further contribute to new and improved solutions. The transport, whether on sea, air or railway, have similar needs and challenges with respect to communication, information, management, planning and costs.

Furthermore the main {\bf MMTS specifications} of the ARKTRANS frameworks that could be a guide for other similar frameworks.
\begin{itemize}
\item[-] A reference model with detailed sub-domains and the roles of the stakeholders;
\item[-] A functional view with detailed functionality of sub-domains;
\item[-] A behavior view with detailed scenarios \& interactions between sub-domains; 
\item[-] An information view with detailed models for freight transport \& MMTS route information;
\end{itemize}

The detailed technical aspects conclude this list. A beneficent interaction within all MMTS process leads to an efficient multimodal transport framework. Overall conclusions specify that MMTS is especially suited for long distances; a major MMTS feature is the total travel time; the access, egress and transfer times could be reduced if there will be integrated MMTS, e.g. park and bicycles facilities, and card access on transit systems.

{\bf Other Frameworks.} Frameworks for multimodal transport security and various policy applications are described in details in the book of~\cite{szyliowicz2016multimodal}.  Other related {\it frameworks and security challenges}, for both passengers and freight and security and policy applications around the world are analyzed in the book of~\cite{wiseman2016multimodal}.

\subsection{Passenger Multimodal Transportation}\label{sec:passengers}

\noindent $>$ {\bf Theoretical approaches.}

\smallskip

\indent -- Designs for chains and networks by~\cite{bockstael2003chains}. As the author describes it "The objective of this research is: Develop a design approach for improving inter-organizational multimodal passenger transport systems from a chain perspective". 

The article raises some interesting aspects, like balancing the positive and negative impacts of mobility, a holistic approach for modes of transportation not necessarily reducing the number of kilometers for passengers, but improving the number of vehicle-kilometers, resulting in a more efficient usage of the resources (infrastructure, fuel).

\smallskip

\indent -- {\bf A highly conceptual approach} by~\cite{chiabaut2015evaluation}. It is applied to a very idealized network. The authors aim to combine different transport modes by extending the concept of Macroscopic Fundamental Diagram (MFD) and therefore, the efficiency of the global transportation system can be assessed. This approach can be applied to a wide range of cases. Although it is an idealized analysis, it provides knowledge about how to compute the overall performance of a multimodal transportation network and methods to compare different traffic management strategies. 

\medskip

\subsection{Models \& Solvers.}

\smallskip

\indent  $>$ {\bf Real-life scenarios.}

\smallskip

\indent -- {\bf A multiobjective linear programming model for passenger pre-trip planning in Greece} by~\cite{aifadopoulou2007multiobjective}. As a case study the trips in Greece using public transport, an integrated web based information gateway was studied. The introduced algorithm (with polynomial complexity) computes the compatibility of various modes based on user preferences, respectively intermodal stations, and identifies the feasible paths. It was structured for checking and certificate optimality; validation on how constrains impacts the computational complexity linear was made; it focuses on a decomposition strategy. Hub selection is significant for compatibility and viability of MMTS; it leads to identify parameters in order to increase compatibility of MMTS services and fees.

\medskip
 
\indent -- {\bf A detailed analysis of the Rhein-Ruhr area} by~\cite{schonharting2003towards}. The authors identified the {\it Rhein-Ruhr area} as a network of corridors (or mega-corridor). Good practices are featured and analyzed with the aim of putting the {\it Rhein-Ruhr area} on the "map" of good examples to follow. 

\medskip

\indent -- {\bf A "waiting time model": case study Tunisian Great Sahel} by~\cite{bouzir2014modeling}. The model  based on multiple variables, was developed in order to optimize waiting times in stations. A case study was based on a survey in the Tunisian Great Sahel. Multiple Correspondence Analysis (MCA) and the General Linear Model were technically used. The new model depends on the following features: the travel cost, purpose and frequencies while using MMTS, and is based on the age of respondents. 

\medskip

The main results of the case study follows.
 \begin{itemize}
\item[-] The MMTS combination including bus \& tram need a longer waiting time than other transportation modes;
\smallskip
\item[-] Young people waits longer for transport services; they use public transport more often than workers; vacationers wait more than daily passengers;
\smallskip
\item[-] MMTS trips including waiting time of taxis is shorter when two transport services are included;
\newpage
\item[-] The semi-collective transportation seems beneficent as reduces waiting time; the semi-public transport with just an transportation mode e.g. taxi cancels reduced waiting time;
\item[-] The travel cost has a major influence in the overall waiting time.
\end{itemize}
The waiting time within public transportation has a direct consequence in the quality of the transportation service. 

\smallskip

\indent -- {\bf TRANSFER model-multimodal network in large cities} by~\cite{carlier2005transfer}. The model, was introduced for analysis of the multimodal network in large cities, as well as route generation. Building Park and Ride (P\&R) keeps automobiles outside city center. More P\&R locations are planned for car drivers as they could park here and further transfer to public transport to further arrive in city centre. The main advantage is making public and/or alternative transport more appealing to passengers.

As any other model, it could be successful if MMTS become more attractive than unimodal transport e.g.car-only trip. The access and egress are also quantified. Here, MMTS are represented as supernetworks where unimodal networks are interconnected by transfer links, the possibility of transfer and related time and costs. TRANSFER components include the following:
\begin{itemize}
\item[-] A multimodal route-set generation module based on network features and passengers preferences;  
\item[-] An assignment module to distribute transport flows among routes; 
\item[-] A path-size route-choice algorithm to avoid overlap among the routes in a route set. 
\end{itemize}
The superbuilder tool was developed, combining some unimodal networks \& transfer data in order to generate a multimodal supernetwork with features of unimodal networks and most relevant transfer possibilities.

\indent-{\bf Transfer points with specific features}  by~\cite{sun2015characterizing}. The authors analyzes age-related transfer speed, the effect of the time of day, the effect of a single person in relation to others, crowding and the use of smart cards. The authors detailed the following:
passenger behavior related to transfer between MTTS modes; correctness of data in order to make a feasible model for passenger transport when complex real-world configurations are provided; efficiently use Smart card data within MMTS. An overall conclusion includes the fact that passengers are faster in the morning no matter if it is crowded or not; children and seniors transfer slower than adults but children outperforming adults through overpasses; further models will have to support pedestrian behaviors and convenient facility design.
\subsection{Freight Multimodal Transportation}\label{sec:freight}
\smallskip

\indent  $>${\bf Real-life scenarios.}

\smallskip

\indent -- {\bf A case study for least-developed economies} where different problems arise, is presented by ~\cite{islam2006promoting} where the situation of Bangladesh is explored from the infrastructure point of view, as well as local bureaucracy. 

In order to evaluate an extent of integration of seaport container terminals in supply chains~\cite{panayides2008evaluating} define and develop specific measures. Optimizing the integration of said container terminals can improve the flow of freight, limiting time waste and delays. 

\medskip

\indent -- {\bf A case study: shipments} focusing on a major iron and steel manufacturer from NW Australia and it's iron ore shipments to NE China, is presented by~\cite{potter2011multimodal}. They studied multiple routes and transport options and even punctual optimization (like congested traffic at a specific moment). Their studies suggest that for long shipments, port variations and inland transport variations have only marginal overall differences, so several combinations of transport and handling methods may successfully coexist. An counter-intuitive conclusion is that is the control of just a company for the entire supply chain, as the bulk cargo market is subject to frequent changes of the prices under global economic conditions.

\medskip

{\bf Others.} In~\cite{yuen2017barriers} Supply Chain Integration with barriers for the maritime logistics industry is discussed. The authors identified a list of barriers from interviews and literature reviews, but also from 172 surveys sent to container shipping companies. There were also identified five factors that cause most of these barrier. Collaborations are also discussed by~\cite{stank2001supply}. An integrated mathematical model of optimal location for transshipment facility in a single source-destination vessel scheduling and transportation-inventory problem was proposed by \cite{al-informatica}. The authors hybrid proposal find a set of cost-effective facility locations and using these locations reduce costs (e.g. daily vessels operations, chartering and penalties costs).

\section{Multimodal Transport Planning }\label{sec:planning}
  
Multiple facets of planning multimodal transport exists, making it more difficult. For example, in a large city, somebody might suddenly decide to engage in a long distance travel. This implies an \textit{ad-hoc} computation of the route and means of transport to be used, according to the individual personal preferences, e.g. not using metro system due to motion-sickness. Planning such a transport means using any available means \textit{at the specific time}; the factors to consider could include: time, cost, weather, waiting times in hubs, etc. What implies planning a diverse transport system for a large city? The designer must compute the available resources, the requirements and even the schedules / working hours of different companies. 

In the planning phase, the designer could suggest the transport means  (buses, trams, etc) in order to obtain an economic and eco-friendly system. An Introduction in Multimodal Transportation Planning book was published by~\cite{litman2017introduction} in which he summarizes the basic principles for multimodal transportation planning for people. He studies transport options for pedestrians, like sidewalk design, bicycles, ride-sharing and public transit systems. He also has very good explanations for multimodal transport planning process, impacts to be considered and are often overlooked and different traffic models, like the Four-Step Traffic Model.
\newpage
\noindent The first stage of planning a multimodal transport system is to understand it's complexities. An complete and accurate model has to be created and analyzed. 

\smallskip

\subsection{Models \& Solvers.}

\smallskip

\noindent $>$ {\bf Planning MMTS with uncertainties and limitations.}

\smallskip

\indent -- {\bf Fuzzy cross-efficiency Data Envelopment Analysis} by~\cite{dotoli2016technique}. The planning of efficient multimodal transports using a fuzzy cross-efficiency Data Envelopment Analysis technique is presented by~\cite{dotoli2016technique}. Other approaches like uncertainty conditions with complex traits and high discriminative power are described. They prove the effectiveness of their approach while studying the optimal transport planning and computing the boundaries of the multi-modal transport. In~\cite{sumalee2011stochastic} the multi-modal transport network with demand uncertainties and adverse weather condition includes formulation of the fixed point problem; other future and existing related development include works of ~\cite{cticalua2017approximating} and~\cite{xu2009multi}.

\indent -- {\bf Metaheuristics for real-time decisions} by ~\cite{mutlu2017planning}. Planning part of multimodal transport with various limitations are reviewed by~\cite{mutlu2017planning}. They discuss problems like real-time decisions in the context of short-term planning, restructuring and re-configuring logistic strategies, and collaborative planning. Appropriate solution methods and intuitive meta-heuristic approaches to rapidly act upon changes are suggested.

\noindent $>$ {\bf Passenger \& freight flows Planning Multimodal transport} 

\smallskip

\indent -- {\bf Multi-Agents Systems for MMTS planning} by ~\cite{greulich2013agent}. As the name suggests, the implementation uses intelligent agents representing various stakeholders and considers the effects of passenger's behavior. 

\indent -- {\bf Genetic Local Search tested in the Java Island, Indonesia} by~\cite{yamada2007optimal}. The research revealed that a procedure based on {\it Genetic Local Search} outperforms in order to find the best combination of alternatives. 

\smallskip

\noindent $>$ {\bf Multi-modal systems.}

\indent -- {\bf A multimodal travel system} by~\cite{bielli2006object}. The authors focused on the network object modeling. This enables the use of the model for computing a shortest path while also integrating multimodal options. They also implement and test a solution for the problem of long-run planning in such systems. 

\indent -- {\bf Syncromodal Transport Planning} by~\cite{mes2016synchromodal}. It is a multimodal planning where the best possible combination of transport modes is selected for each package, is discussed in depth. The syncromodal algorithm is implemented in a 4PL service provider in the Netherlands and managed to obtain a 10.1\% cost reduction and a 14.2\% reduction in $CO_{2}$.

\indent -- {\bf A multimodal transport path sequence: AND/OR graphs facilitate planning} by~\cite{wang2020modeling}. It is proposed a triple-phase generate route method for a feasible multimodal transport path sequence, based on AND/OR graphs. Energy consumption evaluates the multimodal transport energy efficiency. A biobjective optimization model for both energy consumption and route risk is solved with an ant-based technique. 

\noindent The research is limited by the graph complexity; the simulation shows valid and promising results.

\medskip

\noindent $>$ {\bf Traffic Flow Risk Analysis and Predictions.}

\smallskip

\indent -- {\bf Fuzziness approach for risk analysis} by~\cite{stankovic2020new}. A fuzzy Measurement Alternatives and Ranking according to the Compromise Solution, fuzzy MARCOS for Road Traffic Risk Analysis was proposed. The method defines reference points, determined relationships between alternatives \& fuzzy ideal/anti-ideal values and defined utility degree of alternatives in relation to the fuzzy ideal and fuzzy anti-ideal solutions. A case study on a road network of 7.4 km was made. The method supported multi-criteria decision-making, within uncertain environments and its results, in terms of risk, could be further used for improving road safety. Other similar efficient method used to cope with multi-criteria optimization is in by~\cite{multi-g-informatica}. As a plus, parallel processes, e.g.~\cite{parallel-g-informatica}, are effective to optimize objective functions.

\indent -- {\bf A Best-worst method \& triangular fuzzy sets} by~\cite{moslem2020integrated}. It is used for ranking and prioritizing  critical driver uncertain behavior criteria for road safety was studied. The case study uses data from Budapest city: on how drivers perceived road safety issues.

\indent -- {\bf Intelligent transportation system-Bird Swarm Optimizer} by~\cite{zhang2020short}. It includes an Improved Bird Swarm Optimizer used to predict traffic flows; the prediction results are evaluated and  accurate prediction is obtained; the model has positive significance to prevent urban traffic congestion.

\medskip

\section{Optimization of Multimodal Transport}\label{sec:optimization}
 
\medskip
{\par A distributed approach for time-dependant transport networks integrated in the multimodal transport service of the European Carlink platform and validated in real scenarios, was proposed by~\cite{galvez2009distributed}. A real-life validation is included for a specific route from a Belgian city Arlon, to Luxembourg.}

\smallskip

\noindent In the related mobile application implementation within MTS of the Carlink Platform, the requests are sent to the MTS and the users get the shortest path between two selected locations. A framework for selecting an optimal multi-modal route was designed by~\cite{kengpol2014development} based on a multimodal transport cost-model, $CO_{2}$ emissions and even the integrated quantitative risk assessment. This complex optimization targets to minimize transportation costs, transportation time, risk and $CO_{2}$ emission all at once. Multi-node, Multi-mode, Multi-path Integrated Optimization Problems using Hybrid heuristics in the work of~\cite{kang2010research} are studied. They propose an integrated {\it Particle Swarm Optimization (PSO)-Ant Colony Optimization (ACO)} double-layer optimization algorithm.

\smallskip

Hierarchical network structures of transport networks and how the main mechanisms lead to these network structures are the main interests of~\cite{Nes2002DESIGNOM} work.  Optimizing Containerized Transport across multiple choice Multimodal Networks using Dynamic Programming was proposed and successfully tested on a real problem by~\cite{hao2016optimization}. 
\newpage
Route Optimisation Problem using Genetic Algorithms (GA) were proposed by~\cite{jing2012hybrid}; the same technique was used to solve a Multi-Objective Transport System by~\cite{KhanATN19} and could be further extended for the multimodal transport. GA was also used to optimize the time for container handling/transfer, respectively the time at the port by speeding up handling operations by~\cite{kozan1999genetic}.
\subsection{Time Optimization of Multimodal Transport}\label{sec:timeopt}
\medskip

\smallskip

\indent -- {\bf Running time} \& {\bf Rescheduling; solver: Ant Colony Optimization.} \cite{zidi2006ant} proposes an Ant-Colony Optimization (ACO) approach for the rescheduling of multimodal transport networks. The ant-colony approach is best in this case (rescheduling) as it is able to work from a given state and only adapt the solution to the new conditions. Rescheduling is a must, as the system is subject to disturbances (traffic jams, collisions, strikes) which cannot be accounted for at the beginning of the transport, but are very likely to introduce delays or other discrepancies. Furthermore, \cite{zidi2006real}, plans the public transportation system, by using the ant-colony optimization when the theoretical schedule cannot be followed; this approach overcomes the inherent overloading with information of the operators when some problematic situation occurs.

\smallskip

\indent -- {\bf Transport time between nodes; solver: Genetic Algorithms} \& {\it K-sortest path.} \cite{yong2009research} take into consideration the transport time between nodes, time needed for mode change and possible delays. They also present a model that aims to minimize transport and transfer costs, build on a GA based on K-shortest-paths.

\indent -- {\bf Real-time system.} We cannot discuss time optimizations without including~\cite{bock2010real} article about "real-time control of freight forwarder transportation networks". His approach integrates multimodal transportation and multiple transshipments. The real-time system is continually optimized in order to adapt it to the current status of the live data. 
\subsection{Cost Optimization of Multimodal Transport}\label{sec:costopt}
From the perspective of the multimodal logistics provider, cost may be the second most important aspect, immediately after customer satisfaction. This is why cost optimization is one of the concerns of every CEO. 

\smallskip

\indent -- {\bf Cost Optimization with specific criteria using Mixed Integer Linear Programming.} by \cite{sitek2012cost}. The authors included a mathematical model of a multilevel cost-optimization by {\it Mixed Integer Linear Programming (MILP)}. They analyze and integrate in their algorithm, as optimization criteria, factors such as costs of: {\it Production, Transport, Distribution} and {\it Environmental Protection}. Furthermore, all these multiple factors are used by~\cite{sitek2012cost} as optimization criteria into the MILP algorithm, where more criteria are included: timing, volume and capacity. The tests for showing the possibilities of practical decision support and optimization of the supply chain have been performed on sample data.

\indent -- {\bf Cost Optimization including emissions} \& {\bf economies of terminal using Genetic Algorithms} by~\cite{zhang2013optimization}. The authors discussed about environmental costs and introduced a modelling optimization approach for terminal networks, integrating the costs of CO$_{2}$ emissions and economies of terminals. Their proposed algorithm is composed of two levels: the upper level uses genetic algorithms to search for the optimal terminal network configurations; the lower level performs multi-commodity flow assignment over a multimodal network. This model is applied to the Dutch container terminal network.

\smallskip

\subsection{Network Planning and Optimization of Multimodal Transport}\label{sec:netopt}
 
\smallskip

\indent -- {\bf Multiple means into a multimodal system} by~\cite{Nes2002DESIGNOM}. It underlines that a change is needed in today's transportation system, in order to address problems like accessibility of city centres, traffic congestion, but most of all the environmental impact. In this regard, combining multiple means into a truly multimodal system has the ability to capitalize on each subsystem's strengths and limit their weaknesses. Negative factors such as the obligation to transfer, although not very pleasant for the passengers, can have many long-term economical and environmental benefits. So, high quality travel information is crucial.

\indent -- {\bf Abstract perspective of multimodal transport network system} by~\cite{zhang2011multimodal}. Here it is shown the necessity of seamless multimodal traveler information systems; therefore a multimodal transport network system and a test for the model in a study for the Eindhoven region was included

\vspace{0.10cm}

\indent -- {\bf Environmental impact constraint when planning} by~\cite{zhang2013optimization}. and economic development are the two reasons why~\cite{yamada2009designing} say that is crucial to develop and design efficient multimodal networks. They employ a heuristic approach for a complex algorithm with road transport, sea links and freight terminals. The model is successfully applied in a network planning in the Philippines.

\vspace{0.10cm}

\indent -- {\bf Supernetwork equilibrium for supply chain–multimodal transport} by~\cite{yamada2015freight}. A 2 level approach using particle swarm optimization is presented. The upper lavel is solved using particle swarm optimisation, while the lower-level decision use a supply chain–multimodal transport supernetwork equilibrium.
 
\vspace{0.10cm}

 \indent -- {\bf Emergencies solved with an immune affinity model} by~\cite{hu2011container}. The paper proposes a transportation scheduling approach based on immune affinity model. The paper concludes that container multimodal transportation will play an important role in emergency relief, due to the exploitation of the different system's strengths. 

 \vspace{0.10cm}

\indent -- {\bf Practical traffic assignment model for a multimodal transport system with low-mobility groups} by~\cite{zhang2020practical}. Here a route choice equilibrium for specific vehicles and non-vehicles travel times at intersections design is proposed  Validation and verification is made on a case study: Wenling city from China. Some limitations of the models includes ignoring modal choice equilibrium, uncertainty of travel and missing a detailed analysis due to insufficient data.

\vspace{0.10cm}

\indent -- {\bf Bayesian model for transport options} by~\cite{arentze2013adaptive}. A Bayesian Method to learn user preferences and to provide in short time personalized advice regarding transport options is presented; a new sequential attributes processing and an efficient parameter sampling is provided.

\vspace{0.10cm}

\indent -- {\bf Limit cruising-for-parking constraint when planning} by~\cite{zheng2016modeling}. It aims to limit cruising-for-parking; the model is based on the {\it Macroscopic Fundamental Diagram (MFD)} for both single and bi-modal, car and bus in order to reduce costs.

\section{Future possible extensions from Unimodal to Multimodal Transport}\label{sec:theoretical}
As transportation quickly expands worldwide, some existing unimodal transport problems and their solvers could be furthermore extended while including specific requirement to solve multimodal transport problems. Some of these problems are further briefly described.

\medskip

\noindent{\bf A. Supply Chain Networks.} One of the  two-stage supply chain network is considered here to optimize the cost from a manufacturer, to a given number of customers while using a set of distribution centers.

\smallskip

{\bf Models \& Solvers: Supply chain for further Multimodal extension.}
\begin{itemize}
 \item[-] {\bf Multi-Objective Goal Programming} by \cite{roy2017multi}. The mathematical model of Two-Stage Multi-Objective Transportation Problem (MOTP) with the use of a utility function for selecting the goals of the objective functions and numeric tests are included; real-world uncertainty with the use grey parameters (reduced to numbers) are also involved. As for the metrics within objective functions, usually Euclidean distances are used but today for urban related ITS for example could be a plus to use city-block distances as in~\cite{city-informatica} where an evolutionary multimodal optimization technique, with suitable parameters obtains better results than existing techniques.

\smallskip

\item[-] {\bf Genetic Algorithm.} \cite{pop2016hybrid} proposes a heuristic-genetic approach with a hybrid based GA for capacitated fixed-charge problem. Their algorithm was tested on benchmark instances and found to obtain competitive results with other state-of-the-art algorithms. 

\smallskip

\item[-] {\bf Other heuristics.} \cite{chen2017uncertain} studies an {\it Uncertain Bicriteria Solid Transportation problem}; \cite{moreno2016heuristic} employs a heuristic approach for the multiperiod location-transportation problem. Several versions of supply-chain problem including efficient reverse distribution system, secure and green features alongside related solvers are presented by~\cite{pinteasupply2014,pinteasupply2016,pinteasupply2019}. A parallel fast solver where the search domain of solutions is efficiently reduced at each iteration was proposed by~\cite{cosma-informatica} for the two-stage transportation problem with fixed charges. It was identified as a very competitive approach when compared to existing ones with literature dataset.  
\end{itemize}

\newpage
\noindent{\bf B. (Generalized) Vehicle Routing Problem.} For a given set of vehicles and clients, the (G)VRP problem is to determine the optimal set of routes, see~\cite{toth2002vehicle}. This is the one of most studied combinatorial set of problems.\cite{lee-informatica} considers an integrative three-echelon supply chain: Vehicle Routing and Truck Scheduling Problem with a Cross-Docking System; this promising logistics strategy distributes products by eliminating storage and order-picking while using warehouse: directly from inbound to outbound vehicles; a cost optimization EEA-based method was proposed outperforming existing solvers.

Due to its effectiveness many variations of the VRP were built on the basic VRP with extra features, e.g. the Generalized VRP (GVRP). VRP with time windows and VRP pick-up and delivery problems, e.g. solved by~\cite{vrp-informatica} with GA insertion operators, can be further extended to related complex problem. A version of GVRP includes designing optimal delivery or collection routes, subject to capacity restrictions, from a given depot to a number of locations organized in clusters, with the property that exactly one node is visited from each cluster. See~\cite{ghiani2000efficient,pop2013improved} for more details.

{\bf Models \& Solver: (G)VRP for further Multimodal extension.}
\begin{itemize} 
\item[-] Capacitated VRP implies that the vehicles have fixed capacities and the locations have fixed demands in time, see~\cite{toth2002vehicle};
\item[-] VRP \& Multiple Depots involves more depots from which each customer can be served as in~\cite{crevier2007multi}; 		
\item[-] Heterogeneous Fixed Fleet VRP uses a heterogeneous (different types) fleet of vehicles as in~\cite{taillard1999heuristic};

\item[-] Multi-Commodity VRP deals with more commodities per vehicle, which has a set of compartments in which only one commodity can be loaded, the same as in~\cite{repoussis2006hybrid};
\item[-] {\bf Tabu Search and hybridization.} Various heuristic and metaheuristic algorithms have been developed for solving the VRP including: an algorithm based on Tabu Search, adaptive memory and column generation described by~\cite{taillard1999heuristic}. \cite{tarantilis2004threshold}  implemented a threshold accepting procedure where a worse solution is accepted only if it is within a given threshold. A multi-start adaptive memory procedure combined with Path Relinking and a modified Tabu Search was developed by~\cite{li2010adaptive}.  
\item[-] {\bf Iterated Local Search based \& Set Partitioning} \cite{subramanian2012hybrid} described a hybrid algorithm composed by an Iterated Local Search based heuristic and Set Partitioning formulation.
\item[-] {\bf Bio-inspired algorithms.} \cite{matei2015improved} propose an improved immigration memetic algorithm which combines the power of genetic algorithms with the advantages of local search. The article describes the advantages of the immigrational approach on the overall quality of the algorithm (result quality and run-time speed). Diverse versions Genetic Algorithms for solving the current problem are presented by~\cite{matei2010efficient} and \cite{petrovan2019self}. Ant colony methods were used to solve Generalized VRP by~\cite{popGVRPacs2008},~\cite{popdynGVRP2009} and~\cite{pinteaGVRPsens2011}.
\newpage

\item[-] {\bf Heuristics.} In~\cite{leuveano2019integrated} is proposed a heuristic to find optimum inventory replenishment decision when solving transportation \& quality problems into a Just-in-Time (JIT) environment. An vendor-buyer lot-sizing model was proposed; parameters study was included and both capacitated and incapacitated cases were studied. Some advantages of the proposals follows: it obtains feasible solution for inventory replenishment decisions; improve transport payload, reduce defectiveness of products and improves quality-related costs. 
\end{itemize}

\medskip

\noindent {\bf C. (Generalized) Traveling Salesman Problem.} Since 1988 this is one of the most studied problems, and the problem of~\cite{applegate2006traveling} from which the (G)VRP evolved. It is considered a particular case of GVRP when the capacity of the vehicles in infinite, and no intermediary return to the depot is required. Some GTSP versions use a node from each cluster in a route solution, e.g. a city-node from a county-cluster. (G)TSP libraries (see ~\cite{tsplib},~\cite{cooklibdata},~\cite{gtsplib}) are continuously updated mainly based on Geographical Information Systems (GIS) as in~\cite{CrisanCountry2020},~\cite{IberianCrisan2016},~\cite{RomaniaTSP} and~\cite{cooklibcountry}. Integer solvers, e.g.~\cite{neostspsolver}, are feasible for TSP, but solving, large-scale real-life problems requires updated strategies.

\smallskip

{\bf Models \& Solvers (G)TSP for further Multimodal extension.}

\smallskip

\begin{itemize}
\item- Intelligent Transport System - (G)TSP related. In the context of multimodal transportation, the (G)TSP family of problems has many applications; for example could be extend to an related Intelligent Transport System (ITS) as in~\cite{igplITSpintea2020} an~\cite{ITSpintea2018}. Recently, Internet of Things (IoT) was used by~\cite{LuoZZYL19} to connect platforms for ITS.

\smallskip

\item[-] Other - (G)TSP related models. Recent optimization models allow the instances of realistic freight rail transport to be solved, a stage-wise approach for solving the scheduling and routing problems separately nowadays is prevalent. 

\smallskip

\item[-] {\bf Heuristics.} As for VRP, the solvers of (G)TSP are using mainly heuristics: {\it Tabu Search} of~\cite{pedro2013tabu} use local search and accept worsening moves, but introduce restrictions to discourage previously visited solutions; {\it Dynamic Programming}by~\cite{bellman1962dynamic}; {\it Approximation Algorithms}by~\cite{malik2007approximation};

\smallskip

\item[-] {\bf Bio-inspired algorithms} used to solve (G)TSP include: {\it Simulated Annealing}~by~\cite{wang2015solving}; {\it Genetic Algorithms} by~\cite{lin2016solving}~and~\cite{pop2017hybrid} are one of the most straightforward ways to tackle TSP. {\it Ant Colony Optimization (ACO)}used~by~\cite{cheng2007modified,mavrovouniotis2011memetic,dorigo1997ant,pinteasurveygtsp2015} and~\cite{pinteadyngtsp2007} use pheromone trails to optimize routes; a successful interactive Machine Learning (iML) use ACO to solve TSP with the human-in-the-loop approach as in~\cite{appintelliml2019}; some other related natural computing solvers include: {\it Particle Swarm Optimization}~by~\cite{wang2003particle,onwubolu2004optimal}~and~\cite{clerc2004discrete}; {\it Discrete Cuckoo Search Algorithm}~by~\cite{ouaarab2014discrete} is inspired by the breeding behavior of cuckoos using agents' attraction.  
\end{itemize}

\newpage

An {\bf abstract Formalization of Multimodal Transportation} as a concept, is presented by~\cite{ayed2008transfer}. ITS could be expanded by using multi-objective facility location problem models and solvers including heuristics e.g.\cite{firm-informatica}. The graph theory is applied within an algorithm in order to optimize routes and route guidance. The authors try to insert their approach into the Carlink project in order to assess it's performance. \cite{cosma2018hybrid} propose an efficient Hybrid Iterated local Search heuristic procedure to obtain high-quality solutions in reasonable running-time. 

\bigskip

\section{Conclusions}~\label{sec:conclusions}
The current selective survey presents a review of real-world problems, applications and optimization in the integrative multimodal transportation. Transportation is a key element of today's society and a very important engine for economic growth. Some areas (like food, medical supplies) raise transportation to strategic importance, and thus indispensable. 

\medskip

The multimodal transport comes with diverse challenges, e.g. related to security, saving resources and reduce emissions. In the context of today's accelerated global warming, it's more important than ever to do everything to lower pollution as much as possible. An example it is the green multi-modal transport organization approach presented in~\cite{wang2020modelling} where it is validated the China-Europe railway network, reducing the transportation time, increasing energy conservation and lowering carbon emissions, by 40\%, when compared with the classical unimodal water transport.

\medskip

Uncertainties will coexists with multimodal  transportation problem, and as recently~\cite{sharma2020soft} research shows, while using road, rail and air transportation, could be used for example soft sets to model these uncertainties related to the transportation attributes (cost, distance and duration of transport). Multi-criteria shortest path optimization, including time, travel cost and route length, for the NP-complete bus routing problem as in~\cite{bus-informatica} could be further extended for complex ITS problems.

\medskip

Multimodal transportation research is a nowadays challenge and continues, on both theory (e.g. solving complex vehicle routing problem) and applicability, in order to obtain feasible models and solvers for various transportation means on complex conditions, with general and specific attributes. 

\medskip

\subsection*{Acknowledgement}
This work has received funding from the CHIST-ERA BDSI BIG-SMART-LOG and UEFISCDI COFUND-CHIST-ERA-BIG-SMART-LOG Agreement no. 100/01.06.2019.
\newpage
\bibliographystyle{IEEEtran}
\bibliography{file}
\end{document}